# Unleashing the Power of Simplicity: A Minimalist Strategy for State-of-the-Art Fingerprint Enhancement

**Raffaele Cappelli**
Department of Computer Science and Engineering of the University of Bologna, Italy

e-mail: raffaele.cappelli@unibo.it

**ABSTRACT** Fingerprint recognition systems, which rely on the unique characteristics of human fingerprints, are essential in modern security and verification applications. Accurate minutiae extraction, a critical step in these systems, depends on the quality of fingerprint images. Despite recent improvements in fingerprint enhancement techniques, state-of-the-art methods often struggle with low-quality fingerprints and can be computationally demanding. This paper presents a minimalist approach to fingerprint enhancement, prioritizing simplicity and effectiveness. Two novel methods are introduced: a contextual filtering method and a learning-based method. These techniques consistently outperform complex state-of-the-art methods, producing clearer, more accurate, and less noisy images. The effectiveness of these methods is validated using a challenging latent fingerprint database. The open-source implementation of these techniques not only fosters reproducibility but also encourages further advancements in the field. The findings underscore the importance of simplicity in achieving high-quality fingerprint enhancement and suggest that future research should balance complexity and practical benefits.

## I. INTRODUCTION

Fingerprint recognition systems have become essential tools in modern security and verification applications, capitalizing on the inherent uniqueness and reliability of human fingerprints [1]. A fingerprint is an imprint of the friction *ridges* and *valleys* on the epidermis of a fingertip (Fig. 1). *Minutiae*, including *ridge endings* and *bifurcations*, are key features used in fingerprint matching since they are distinctive and permanent pattern disruptions.

Image quality directly affects the performance of fingerprint recognition systems, especially in the accurate extraction of minutiae [1]. Low-quality images often result in errors during minutiae detection, thereby compromising accuracy and reliability [2]. For decades, scientific research has focused on developing fingerprint enhancement techniques to improve the quality of degraded fingerprints, such as latent prints. From the beginning of fingerprint enhancement research, two image features — local ridge orientations and frequencies (Fig. 1) — have been considered fundamental for achieving effective results. Early methods primarily relied on *contextual filtering*, where the local context was defined by these features [1]. The advent of deep learning in the past decade has led to a surge in methods leveraging neural networks, particularly Convolutional Neural Networks (CNNs) and Generative Adversarial Networks (GANs). Although these recent methods have led to improvements in fingerprint enhancement, they still fall short of delivering satisfactory performance on low-quality latent fingerprints; moreover, their complexity is quite high and often hinders practical implementation and usability.

Inspired by the KISS (Keep It Simple and Straightforward[1]) principle [3] [4], this work demonstrates that state-of-the-art performance can be achieved using extremely simple and linear methods. The main contributions of this paper are as follows.

- *Outperforming complex methods*: A simple contextual filtering approach is shown to surpass the performance of state-of-the-art methods. On a challenging database of latent fingerprints, this method allows to improve minutiae extraction from enhanced images compared to existing techniques in terms of $F_1$-*score*.

---

[1] While other interpretations of the KISS acronym are more common, this paper employs a less colloquial but more academically appropriate definition in the interest of maintaining scientific precision and avoiding potential misinterpretations.

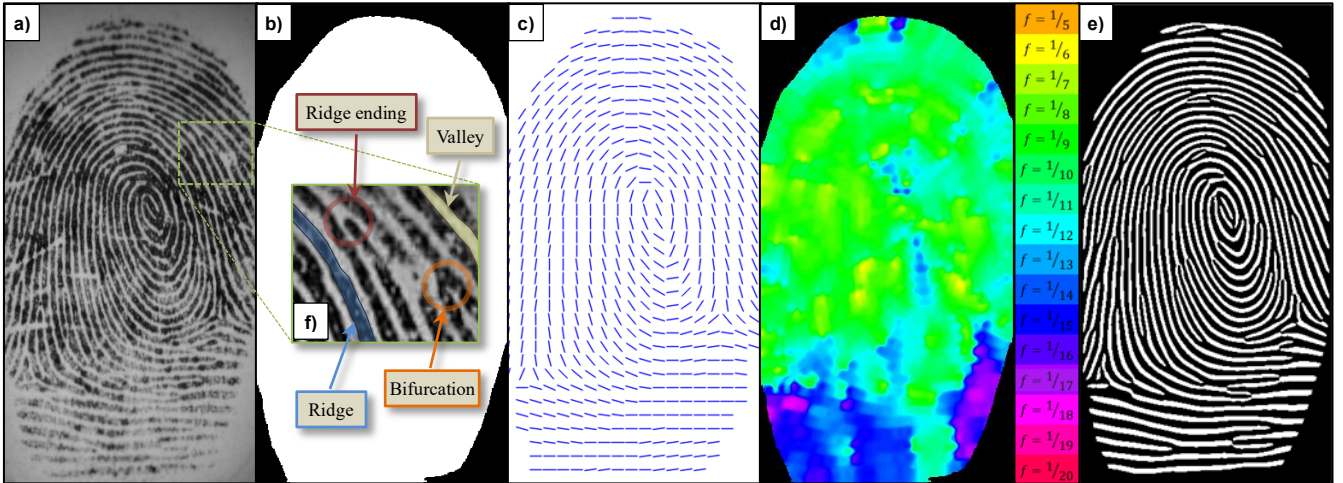

**FIGURE 1.** Example of a fingerprint, its features, and its corresponding enhanced image: (a) original fingerprint image, (b) segmentation mask, (c) orientation field (downsampled to $1/256^{th}$ of its original resolution, with line segments indicating local orientations every 16 pixels), (d) frequency map (visualized as a color-coded image), (e) enhanced fingerprint image (binary image with white ridges and black valleys), and (f) magnified region highlighting a ridge, a valley, and two minutiae (a ridge ending and a bifurcation).

- *Introducing a novel, simple learning-based method*: A new learning-based approach is proposed that not only maintains simplicity, linearity, and ease of implementation but also achieves an even higher $F_1$-*score* than the first method. This approach features a simple network architecture that can be trained in just 25 minutes, in contrast to other state-of-the-art methods which require hours.
- *Promoting reproducibility*: An open-source implementation of both methods is provided[2], to facilitate future research and experimentation in this field.

The remainder of this paper is organized as follows. Section II reviews the main fingerprint enhancement methods proposed in the literature. Sections III and IV describe the two novel enhancement methods, respectively. Section V reports experiments aimed at evaluating the performance of the proposed methods and comparing them to the state-of-the-art on a challenging benchmark of latent fingerprints. Finally, Section VI draws some concluding remarks.

## II. RELATED WORKS

Table 1 summarizes the primary fingerprint enhancement methods published over nearly four decades.

Most pre-deep-learning methods employed contextual filtering. Unlike standard convolution, which applies a single filter to all image pixels, contextual filtering selects a distinct filter for each pixel based on features extracted from its neighborhood. For fingerprints, these features primarily involve ridge orientation and frequency. Only a few pre-deep-learning papers ([5], [6], [7]) deviated from this contextual filtering approach. These studies applied all filters from a predefined bank selecting the filter whose response was the most suited according to a given metric (see the "Contextual filtering" column).

As shown in Table 1, most methods analyze fingerprints in the spatial domain; approximately one-quarter operate in the frequency domain (see the "Frequency domain" column).

The vast majority of convolution-based methods use Gabor filters [8] because of their ability to select and analyze specific orientations and frequencies. Some studies proposed variations on traditional Gabor filters (see the "Modified Gabor Filters" column in Table 1). For example, [9] employed filters with distinct positive and negative peak periods to accommodate varying ridge/valley widths, [10] suggested replacing Gabor filters with Log-Gabor filters, and [11] proposed curved Gabor filters to better adapt to high-curvature image areas. Similar filters were also used in [12]. Finally, some methods adapt Gabor filter bandwidth based on the coherence of local orientations [13] [14].

More recently, Total Variation-based image decomposition into cartoon and texture components [15] has become a popular preprocessing step for enhancement methods (see the "Cartoon-texture decomposition" column).

Some methods employ iterative approaches (see the "Iterative processing" column) to achieve greater noise robustness, albeit at the cost of increased computational complexity. Examples include orientation diffusion-based methods [16] [17] and techniques that initiate enhancement in high-quality regions and progressively extend to lower-quality regions [7] [18]. Even recent deep learning-based methods sometimes use iterative approaches (e.g., [19] [20] [21] [22]).

Multiresolution analysis is a prevalent approach for mitigating the intricate noise patterns inherent in fingerprint images (see the "Multiresolution" column). For instance, [23] leveraged wavelet decomposition, while [24] employed a Laplacian-like image-scale pyramid. More recently, [20] introduced a method that progressively trains a neural network by gradually increasing the resolution of training images. It is worth noting that most CNNs implicitly incorporate multiresolution analysis through pooling layers or convolutions with strides greater than one (see "Multiresolution" and "Convolutional Neural Network" columns).

---

[2] https://github.com/raffaele-cappelli/pyfing

**TABLE 1.** Summary of fingerprint enhancement methods, categorized by key characteristics.

| Method | Year | Contextual filtering | Gabor filters | Modified Gabor filters | Iterative processing | Multiresolution | Frequency domain | Cartoon-texture decomposition | Learning-based | Convolutional Neural Network | Generative Adversarial Network | Trained on synthetic fingerprints |
|---|---|---|---|---|---|---|---|---|---|---|---|---|
| O'Gorman and Nickerson [25] | 1988 | ✓ | | | | | | | | | | |
| Sherlock, Monro, and Millard [26] | 1994 | ✓ | | | | | ✓ | | | | | |
| Kamei and Mizoguchi [5] | 1995 | | | | | | ✓ | | | | | |
| Hong, Wan, and Jain [27] | 1998 | ✓ | ✓ | | | | | | | | | |
| Erol, Halici, and Ongun [13] | 1999 | ✓ | ✓ | | | | | | | | | |
| Almansa and Lindeberg [28] | 2000 | ✓ | | | ✓ | ✓ | | | | | | |
| Greenberg et al. [29] | 2000 | ✓ | ✓ | | | | | | | | | |
| Willis and Myers [30] | 2001 | ✓ | | | | | ✓ | | | | | |
| Hsieh, Lai, and Wang [23] | 2003 | ✓ | | | | ✓ | | | | | | |
| Yang et al. [9] | 2003 | ✓ | ✓ | ✓ | | | | | | | | |
| Nakamura et al. [6] | 2004 | | ✓ | | | | | | | | | |
| Chikkerur, Govindaraju, and Cartwright [31] | 2005 | ✓ | | | | | ✓ | | | | | |
| Wu and Govindaraju [14] | 2006 | ✓ | ✓ | | | | | | | | | |
| Jirachaweng and Areekul [32] | 2007 | ✓ | | | | | ✓ | | | | | |
| Fronthaler, Kollreider, and Bigun [24] | 2008 | ✓ | | | | ✓ | | | | | | |
| Wang et al. [10] | 2008 | ✓ | ✓ | ✓ | | | | | | | | |
| Zhao et al. [16] | 2009 | ✓ | | | ✓ | | | | | | | |
| Rama and Namboodiri [33] | 2011 | | | | | ✓ | | | ✓ | | | |
| Turroni, Cappelli, and Maltoni [7] | 2012 | | ✓ | | ✓ | | | | | | | |
| Gottschlich [11] | 2012 | ✓ | ✓ | ✓ | | | | | | | | |
| Gottschlich and Schönlieb [17] | 2012 | ✓ | | | ✓ | | | | | | | |
| Sutthiwichaiporn and Areekul [18] | 2013 | ✓ | | | ✓ | ✓ | ✓ | | | | | |
| Feng, Zhou, and Jain [34] | 2013 | ✓ | ✓ | | | | | | ✓ | | | |
| Yang, Feng, and Zhou [35] | 2014 | ✓ | ✓ | | | | | | ✓ | | | |
| Cao, Liu, and Jain [36] | 2014 | ✓ | ✓ | | | ✓ | | ✓ | ✓ | | | |
| Schuch, Schulz and Busch [37] | 2016 | | | | | | | | ✓ | ✓ | | ✓ |
| Svoboda, Monti and Bronstein [38] | 2017 | | | | | ✓ | | | ✓ | ✓ | | ✓ |
| Tang et al. [39] | 2017 | ✓ | ✓ | | | ✓ | | | ✓ | ✓ | | |
| Chaidee, Horapong, and Areekul [12] | 2018 | ✓ | ✓ | ✓ | | | ✓ | ✓ | ✓ | | | |
| Dabouei et al. [40] | 2018 | | | | | ✓ | | | ✓ | ✓ | ✓ | |
| Li, Feng and Kuo [41] | 2018 | | | | | ✓ | | | ✓ | ✓ | | |
| Joshi et al. [42] | 2019 | | | | | ✓ | | | ✓ | ✓ | ✓ | ✓ |
| Qian, Li and Liu [19] | 2019 | | | | ✓ | ✓ | | | ✓ | ✓ | | |
| Cao et al. [43] | 2020 | ✓ | ✓ | | | ✓ | ✓ | ✓ | ✓ | ✓ | | |
| Huang, Qian and Liu [20] | 2020 | | | | ✓ | ✓ | | | ✓ | ✓ | ✓ | |
| Wong and Lai [44] | 2020 | | | | | | | | ✓ | ✓ | | ✓ |
| Horapong, Srisutheenon, and Areekul [21] | 2021 | | | | ✓ | | ✓ | ✓ | ✓ | ✓ | | |
| Liu and Qian [45] | 2021 | | | | | ✓ | | | ✓ | ✓ | | |
| Zhu, Yin and Hu [46] | 2023 | | | | | ✓ | | ✓ | ✓ | ✓ | ✓ | |
| Kriangkhajorn, Horapong, and Areekul [22] | 2024 | | | | ✓ | ✓ | ✓ | ✓ | ✓ | ✓ | | |
| Pramukha, Akhila, and Shashidhar [47] | 2024 | | | | | ✓ | | | ✓ | ✓ | ✓ | ✓ |

Over the past decade, almost all methods have relied on learning techniques (see the "Learning-based" column in Table 1), principally deep learning models such as CNNs with encoder-decoder architectures (see the "Convolutional Neural Network" column). However, some studies have employed other learning approaches, including a Markov Random Field-based method [33] and four dictionary-based methods, three in the spatial domain [34] [35] [36] and one in the frequency domain [12]. While deep learning yielded improved results, it also introduced the challenge of obtaining large quantities of fingerprints with corresponding ground truth. Some researchers addressed this issue using synthetic fingerprint generators [48] (see the "Trained on synthetic fingerprints" column).

Finally, more recent studies have combined CNNs with GANs (see the "Generative Adversarial Network" column in Table 1) to further improve the results [20] [40] [42] [46] [47].

## III. GBFEN: GABOR FILTERING STRIKES BACK

The year 2024 marked a significant leap forward in fingerprint feature extraction. Notably, [49] introduced a groundbreaking neural network-based approach for orientation field estimation, surpassing all prior methods on standard benchmarks. Additionally, [50] pioneered a technique for accurate ground-truth labeling of local frequencies, enabling the training of a highly effective neural network for frequency estimation. Building upon these advancements, a fundamental question arose: Can these novel techniques be harnessed to significantly improve fingerprint enhancement? To address this, the Gabor-Based Fingerprint ENhancement (GBFEN) method was designed. GBFEN comprises three steps, as depicted in Fig. 2:

1. *Orientation estimation* – Given a fingerprint image **F** and its segmentation mask **S**, a CNN, as described in [49], is employed to estimate the dense orientation field **Θ**, which contains the local orientation $Θ_{x,y}$ at each pixel $(x, y)$. The network architecture, a streamlined encoder-decoder model, culminates in a Head block that translates the internal representation into explicit angles $Θ_{x,y} \in [0, π)$.

2. *Frequency estimation* – A CNN similar to the one used in the previous step (see [50]), estimates the dense frequency map **ℱ**, which contains the local frequency $f_{x,y}$ at each pixel $(x, y)$. To incorporate orientation information, an initial Stem block processes the fingerprint image **F**, its segmentation mask **S**, and the orientation field **Θ** into a four-channel tensor, which is subsequently fed into the encoder.

3. *Contextual filtering* – A bank of Gabor filters is pre-computed. For each pixel $(x, y)$ in fingerprint **F**, the filter closest to orientation $Θ_{x,y}$ and frequency $f_{x,y}$ is selected. Convolution with this filter, applied to a local neighborhood centered at $(x, y)$, yields the corresponding enhanced pixel value. This process is repeated for each pixel, to obtain the final enhanced image **E**.

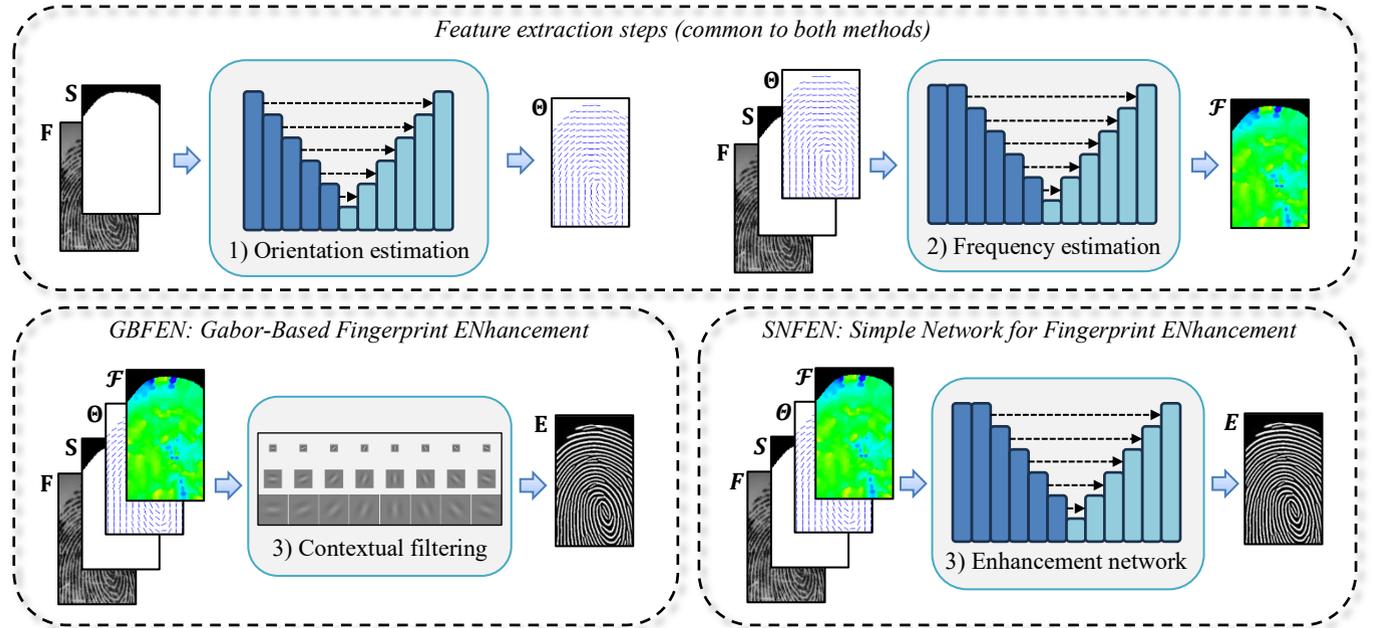

**FIGURE 2. Main processing steps of the two proposed methods.** First, a CNN (see [49]) processes the fingerprint image F and its segmentation mask S to estimate the orientation field Θ. Next, another CNN (see [50]), incorporating the orientation field, estimates the frequency map ℱ. The third step is the actual fingerprint enhancement, which differentiates the two methods: GBFEN applies Gabor filters to F, while SNFEN adopts a simple encoder-decoder architecture (see Fig. 4). In both cases, the resulting filtered image E represents the final enhanced output.

The Gabor filters employed in GBFEN have the following form:

$$g_{\theta,f}(x,y) = e^{-\frac{x_\theta^2 + y_\theta^2}{2\sigma^2}} \cdot \cos(2\pi f x_\theta) \quad (1)$$

where $x_\theta = x \cdot \sin(\theta) + y \cdot \cos(\theta)$, $y_\theta = -x \cdot \cos(\theta) + y \cdot \sin(\theta)$, and $\sigma = \frac{5}{12f}$.

$g_{\theta,f}(x,y)$ is the value, at position $(x,y)$, of the kernel of a Gabor filter with orientation $\theta$ and frequency $f$. Note that, while in previous studies the standard deviation $\sigma$ of the Gaussian envelope was usually fixed (see for instance [27] and [35]), GBFEN adjusts it according to the local frequency $f$ (see Fig. 3). The size $s$ of the square kernel matrix is computed as an integer odd number as follows: $s = 1 + 2 \cdot \lceil 3\sigma \rceil$ (Fig. 3). Each filter is finally standardized to have zero-mean and unit norm. A total of 144 filters (16 different orientations and 9 different frequencies) are used as the pre-computed filter bank in GBFEN.

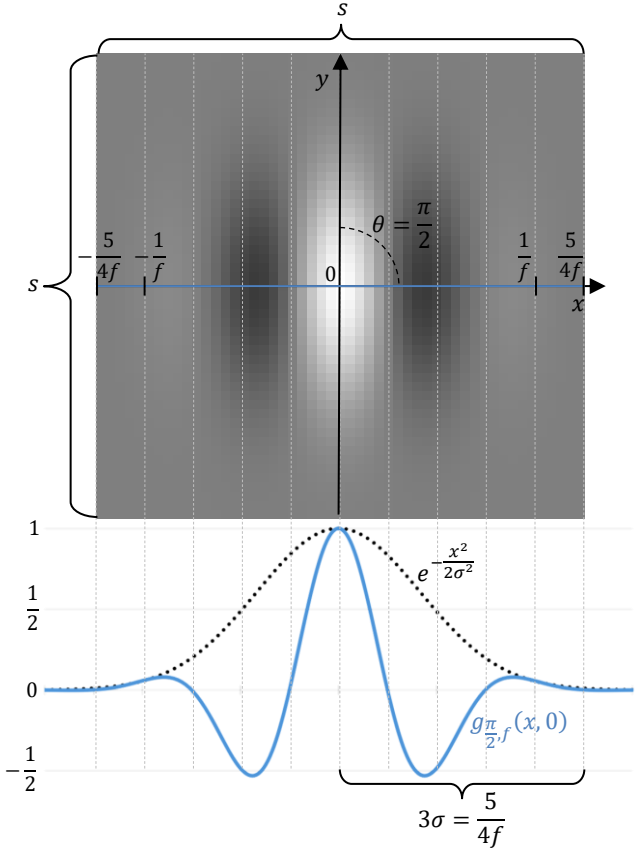

**FIGURE 3.** Example of a Gabor filter with $\theta = \frac{\pi}{2}$ and the relationships between its parameters. The filter is an $s \times s$ square matrix, which is displayed as an image in the upper part of the figure, where lighter pixels correspond to positive values and darker pixels to negative values. The graph in the lower part of the figure shows in blue the filter's cross-section along the $x$-axis and its Gaussian envelope as a dotted black line. The $x$-coordinates corresponding to the filter's period ($f^{-1}$) and to $3\sigma$ are highlighted to better illustrate the defined relationship between $s$, $\sigma$, and $f$.

## IV. SNFEN: A SIMPLE DEEP LEARNING APPROACH

SNFEN (Simple Network for Fingerprint ENhancement) was designed to investigate whether a streamlined CNN, leveraging state-of-the-art features, could surpass the performance of traditional Gabor filters. Fig. 2 illustrates the end-to-end pipeline of the proposed method. The initial two stages align with those of GBFEN, while the third stage introduces a novel CNN architecture specifically trained to produce enhanced fingerprint images from four input modalities: the original fingerprint image **F**, its segmentation mask **S**, its orientation field **Θ**, and its frequency map **𝓕**.

### A. ARCHITECTURE

Fig. 4 depicts the fingerprint enhancement network of SNFEN. The network comprises four key building blocks: the Stem, the Encoder, the Decoder, and the Head. The Stem block transforms the orientation field **Θ**, containing an angle in radians $\mathbf{\Theta}_{i,j} \in [0,\pi]$ for each pixel $(i,j)$, into a double-angle representation (see [49]):

$$\mathbf{X}_{i,j} = \cos(2 \cdot \mathbf{\Theta}_{i,j}), \mathbf{Y}_{i,j} = \sin(2 \cdot \mathbf{\Theta}_{i,j}) \quad (2)$$

The two resulting feature maps are concatenated with **F**, **S**, and **𝓕**, forming a five-channel tensor that is fed into the Encoder.

The Encoder and Decoder blocks have a symmetrical structure, each composed of five levels. Each encoder level incorporates a $5 \times 5$ convolution with padding and ReLU activation, followed by batch normalization and $2 \times 2$ max pooling for downsampling. Conversely, each decoder level employs a $5 \times 5$ convolutional layer with padding and ReLU activation, batch normalization, and a $2 \times 2$ spatial upsampling. To preserve high-resolution details crucial for effective enhancement, skip connections are used. These connections transmit feature maps from the encoder levels to their corresponding decoder levels, where they are concatenated with the upsampled features.

The Head block operates at the same resolution as the input fingerprint. It consists of 16 $5 \times 5$ convolutional filters with padding, ReLU activation, and batch normalization, followed by a final $5 \times 5$ convolutional layer with sigmoid activation to produce the enhanced pixel values.

Fig. 5 provides a visual representation of the network's processing pipeline. From top to bottom, it illustrates: 1) the input modalities (**F**, **S**, **Θ**, and **𝓕**), 2) the output of the Stem block (where **Θ** is transformed into the two feature maps of the double-angle representation), 3) the output of each convolutional layer in the Encoder and Decoder blocks (visualizing only the first 16 feature maps), 4) the output of the first convolutional layer in the Head block, and 5) the final enhanced image **E**. It is important to note that while the feature maps are resized for visualization purposes, their actual resolutions vary from the original fingerprint size to $1/32^{th}$ of the original resolution. As we journey through the encoding path, the feature maps undergo a fascinating transformation. Initially, they capture the intricate details and localized patterns of the input fingerprint, sometimes intertwining with the orientation and frequency information. Gradually, they evolve into representations of higher-level contextual information and global structures, increasingly attuned to the enhancement task guided by the local orientation and frequency cues. Conversely, the decoding path embarks on a refining process, transforming abstract spatial representations into precise pixel-level maps. This is aided by the infusion of high-resolution details channeled through the skip connections. The first convolutional layer of the Head block generates 16 feature maps that bear a striking resemblance to fragments of the final enhanced image. These fragments seem to correspond to specific regions of the fingerprint, possibly related to particular orientation and frequency ranges. The exact nature of this specialization is not yet fully understood, but it suggests that the network has learned to decompose the enhancement task into smaller, more manageable subtasks.

### B. TRAINING DATA AND GROUND TRUTH

Table 2 presents the fingerprint datasets employed for training and preliminary experimentation to select the optimal network architecture, loss function, and hyperparameters. These datasets were sourced from FVC2002 [51], FVC2004 [52] and the FFE benchmark [50]. For each fingerprint used in training, the following ground truth data were assembled (Fig. 6).

- Segmentation mask $S$: this was already available in the FFE benchmark, while for the FVC datasets it was obtained from the manually marked ground truth data provided by Thai, Huckemann, and Gottschlich [53].
- Orientation field $\hat{\Theta}$: A specialized software tool [49] was used to meticulously annotate the orientation field.
- Frequency map $\hat{\mathcal{F}}$: derived from the manually-annotated ridge skeleton $\hat{\xi}$ using the procedure outlined in [50].
- Enhanced fingerprint $\hat{E}$: generated by applying the contextual convolution step of GBFEN to the manually-annotated ridge skeleton $\hat{\xi}$, using the ground truth orientation field $\hat{\Theta}$ and frequency map $\hat{\mathcal{F}}$. The resulting ground-truth enhanced fingerprint is a nearly binary image with white ridges and black valleys (Fig. 6).

**TABLE 2.** Fingerprint datasets used for training and for selecting the model and hyperparameters.

| Dataset | Size | Usage |
|---|---|---|
| FVC2002 DB2 A (3rd impression of each finger) | 100 | Training |
| FVC2002 DB3 A (3rd impression of each finger) | 100 | Training |
| FVC2004 DB1 A (1st impression of each finger) | 100 | Training |
| FFE benchmark ("Good" and "Bad" datasets) | 60 | Training |
| FVC2002 DB1 A (1st impression of each finger) | 100 | Model selection |

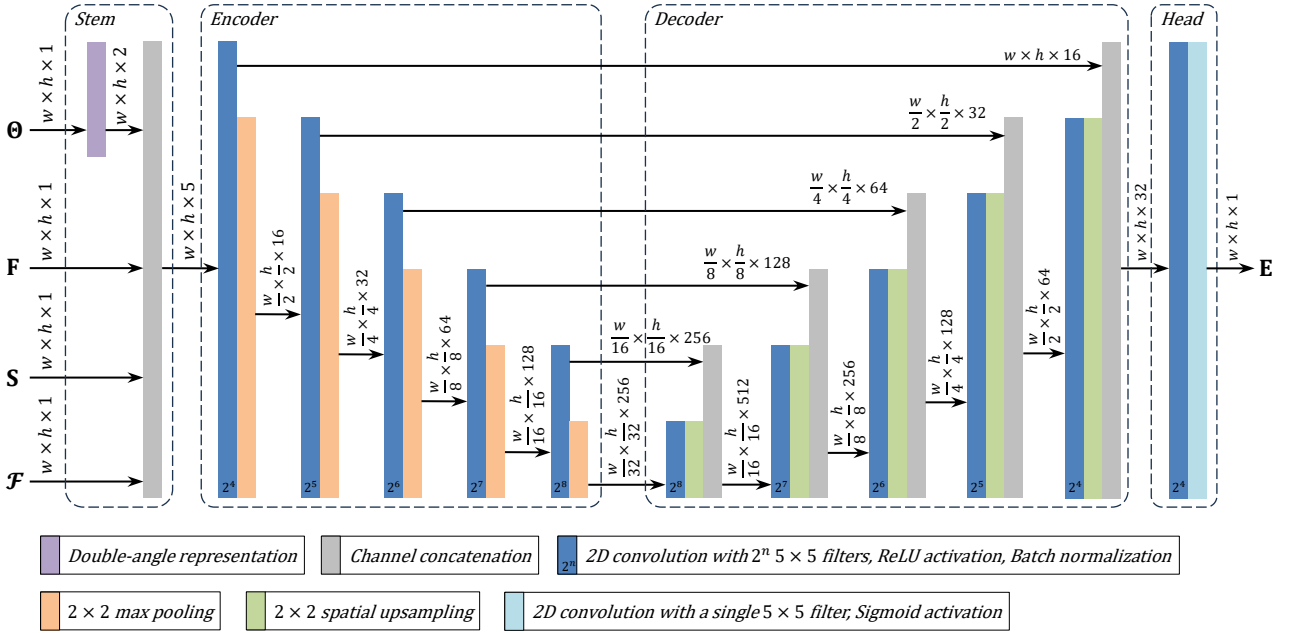

**FIGURE 4.** Architecture of the fingerprint enhancement network in SNFEN. This figure illustrates the input data (fingerprint $F$, segmentation mask $S$, orientation field $\Theta$, and frequency map $\mathcal{F}$), the main building blocks (Stem, Encoder, Decoder, Head), and the resulting enhanced image $E$. The various types of network layers are graphically shown in different colors, according to the legend above: each arrow represents information flow in the form of a multi-channel tensor, whose size is specified near the arrow itself in the form $width \times height \times channels$.

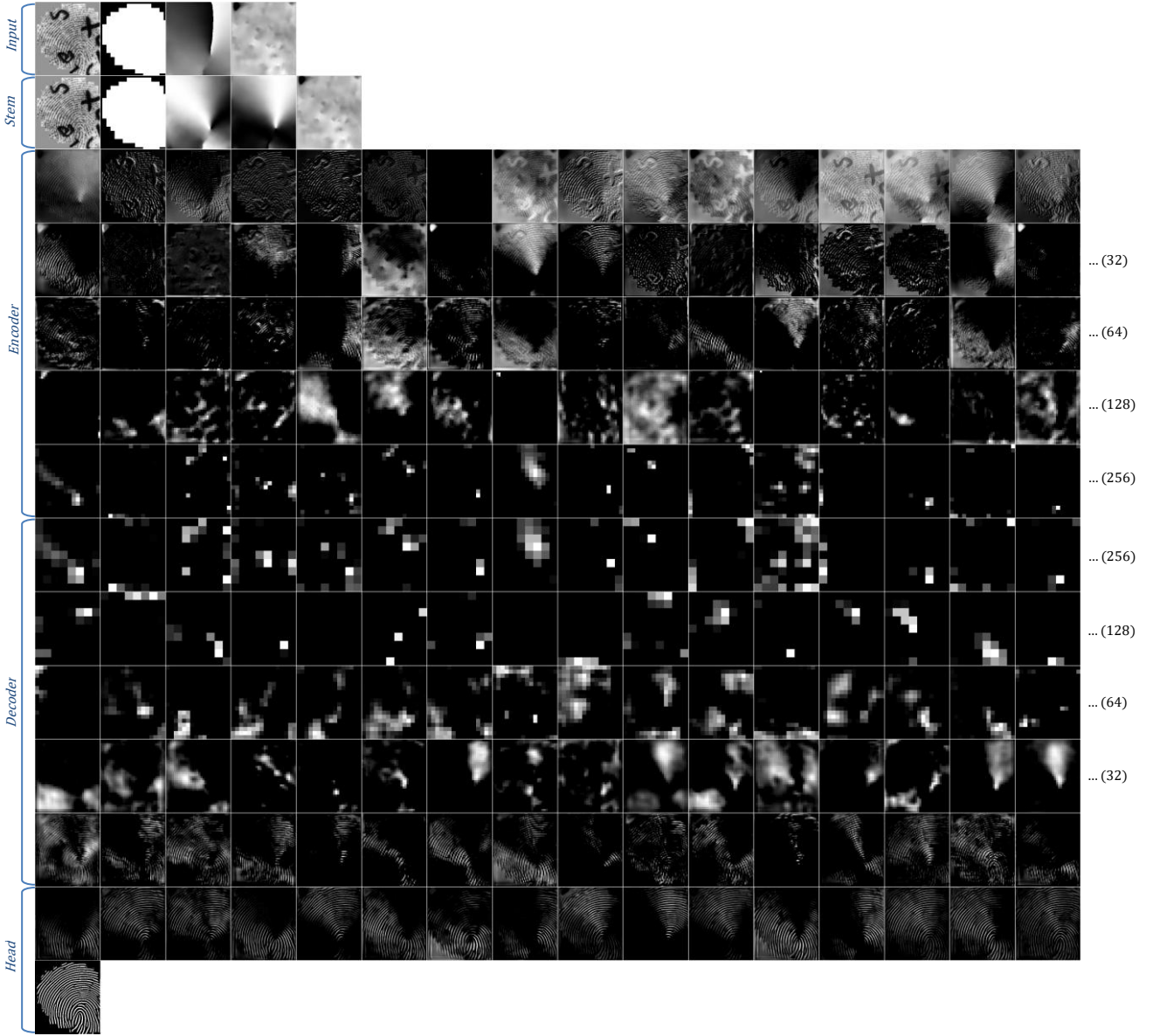

**FIGURE 5.** Example of input, feature maps, and output of the SNFEN enhancement network. The first 16 feature maps produced by each convolutional layer are shown, all resized at the same dimensions for visualization purposes.

## C. LOSS FUNCTION

Network training is guided by a loss function based on the Tversky index [54].

$$\mathcal{L}(\mathbf{E},\hat{\mathbf{E}},\mathbf{S}) = 1 - \frac{TRA(\mathbf{E},\hat{\mathbf{E}},\mathbf{S})}{TRA(\mathbf{E},\hat{\mathbf{E}},\mathbf{S}) + \alpha FRA(\mathbf{E},\hat{\mathbf{E}},\mathbf{S}) + (1-\alpha)FVA(\mathbf{E},\hat{\mathbf{E}},\mathbf{S})} \quad (3)$$

Where:
- $\mathbf{E}$ is the enhanced image predicted by the network, with pixel values $\mathbf{E}_{i,j} \in [0,1]$ representing the probability of a ridge at that location.
- $\hat{\mathbf{E}}$ is the ground truth enhanced fingerprint.
- $\mathbf{S}$ is the ground truth segmentation mask, with $\mathbf{S}_{i,j} = 1$ for foreground pixels, zero otherwise.
- $TRA(\mathbf{E},\hat{\mathbf{E}},\hat{\mathbf{S}}) = \sum_{i,j} \mathbf{E}_{i,j} \cdot \hat{\mathbf{E}}_{i,j} \cdot \mathbf{S}_{i,j}$ is the total agreement between the predicted and ground-truth ridge probabilities within the foreground region (*True Ridge Agreement*).
- $FRA(\mathbf{E},\hat{\mathbf{E}},\hat{\mathbf{S}}) = \sum_{i,j} \mathbf{E}_{i,j} \cdot (1 - \hat{\mathbf{E}}_{i,j}) \cdot \mathbf{S}_{i,j}$ is the total agreement between the predicted ridge probabilities and the ground-truth valley probabilities within the foreground region (*False Ridge Agreement*).
- $FVA(\mathbf{E},\hat{\mathbf{E}},\hat{\mathbf{S}}) = \sum_{i,j} (1 - \mathbf{E}_{i,j}) \cdot \hat{\mathbf{E}}_{i,j} \cdot \mathbf{S}_{i,j}$ is the total agreement between the predicted valley probabilities and the ground truth ridge probabilities within the foreground region (*False Valley Agreement*).
- $\alpha \in [0,1]$ is a parameter controlling the relative weight of $FRA$ errors to $FVA$ errors.

This loss function was selected from a set of candidate loss functions, including Binary Cross Entropy, Dice Loss, and Focal Loss [55], based on preliminary experiments on a disjoint dataset (see Table 2). The value of parameter $\alpha$ was set to 0.7, as suggested in other studies [56] [57].

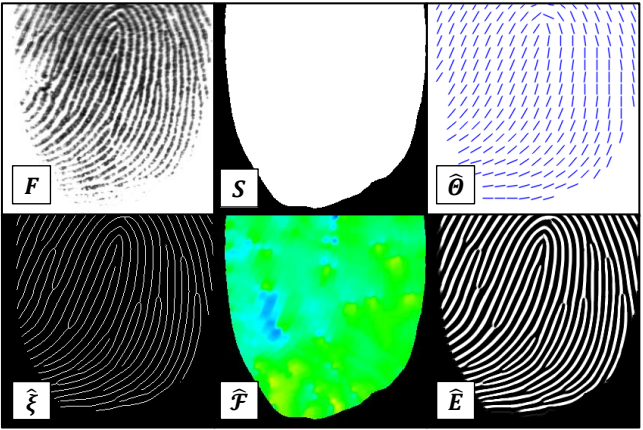

FIGURE 6. Example of a training fingerprint F and its corresponding ground truth data. S: manually labeled segmentation mask. $\hat{\Theta}$: manually labelled orientation field (downsampled to $1/64^{th}$ of its original resolution, with line segments indicating local orientations every 8 pixels). $\hat{\xi}$: manually labeled ridge skeleton. $\hat{\mathcal{F}}$: frequency map (derived from the skeleton and visualized as a color image with the color scale in Fig. 1). $\hat{E}$: enhanced fingerprint (generated from the skeleton $\hat{\xi}$).

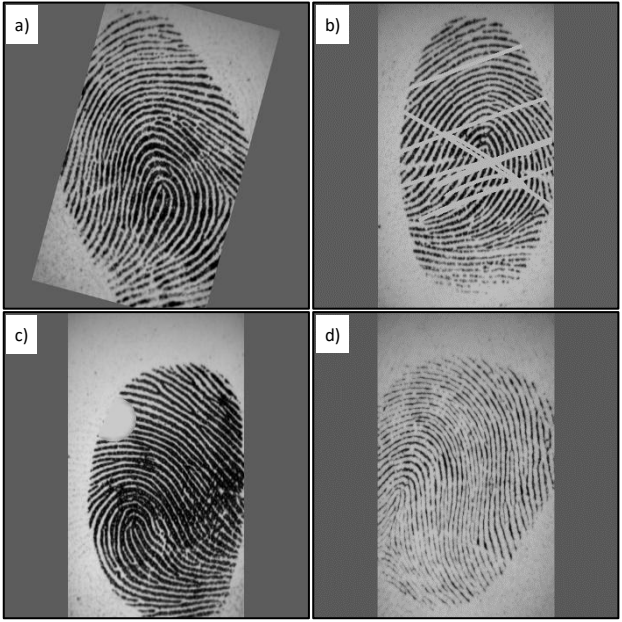

FIGURE 7. Examples of data augmentation applied to training fingerprints: (a) clockwise rotation of 15°, (b) random lines to simulate scratches, (c) an elliptical blob to simulate an abrasion, and (d) morphological dilation, resulting in thicker valleys and thinner ridges.

### D. TRAINING AND HYPERPARAMETERS

A series of exploratory experiments were conducted to identify the most promising training procedure, hyperparameters, and data augmentation strategy, leading to the following setup.

- *Input image size* – all training fingerprints were padded to a uniform size of $512 \times 512$ pixels.
- *Data augmentation* – various fingerprint-specific augmentation techniques were implemented, including random translation ($\pm 5\%$), rotation ($\pm 20°$), scale ($\pm 15\%$), horizontal flip, gamma correction, contrast reduction, morphological operations (erosion or dilation), simulated scratches and abrasions (Fig. 7).
- *Optimization algorithm* – the Lion optimizer [58] was employed. The learning rate was dynamically adjusted between $10^{-7}$ and $10^{-3}$ using a cosine decay strategy with warmup. The momentum parameters were set to $\beta_1 = 0.2$ and $\beta_2 = 0.5$.
- *Training epochs* – Due to the limited availability of fingerprints in the training datasets, a validation set was not used to determine training termination. Instead, a fixed number of 25 training epochs was established, with each epoch consisting of 300 batches of 16 fingerprints.

The entire training process took about 25 minutes on a PC with an NVIDIA GeForce RTX™ 3080 Ti GPU.

## V. EXPERIMENTAL RESULTS

### A. BENCHMARK DATASET

To rigorously evaluate the efficacy of the proposed fingerprint enhancement methods and benchmark them against state-of-the-art techniques, the challenging NIST SD27 latent fingerprint database [59] was employed. Curated by the National Institute of Standards and Technology in collaboration with the FBI, this dataset comprises 258 latent fingerprints meticulously collected from real-world crime scenes. These fingerprints are categorized into three distinct quality levels: "Good", "Bad", and "Ugly", with a balanced distribution of 88, 85, and 85 images, respectively. As illustrated in Fig. 8, the database's wide spectrum of fingerprint quality and degradation, including heavy noise, distortion, and partial fingerprints, arising from various environmental and acquisition factors, makes it a comprehensive evaluation platform for assessing the performance of enhancement algorithms. For each image, the NIST SD27 also contains the minutiae ground truth, validated by a professional team of latent examiners. Ground truth segmentation masks for NIST SD27 images were provided by J. Feng et al. in [34].

### B. IMPLEMENTATION DETAILS

Both GBFEN and SNFEN are implemented in Python, leveraging the capabilities of the OpenCV library [60] for image processing tasks and the Keras library [61] for building and training deep neural networks.

The execution time of the algorithms was evaluated on a workstation equipped with an Intel® Xeon® Silver 4112 CPU operating at 2.60 GHz and an NVIDIA GeForce RTX™ 3080 Ti GPU. When processing fingerprints from the NIST DB27 dataset, GBFEN exhibits an average execution time of 518 milliseconds, with the contextual convolution step contributing approximately 320 milliseconds to the overall processing time. SNFEN achieves a faster average execution time of 301 milliseconds, with the fingerprint enhancement network consuming roughly 103 milliseconds. The execution time of both methods is therefore compatible with use in real-time applications, such as law-enforcement or security systems [1]. Furthermore, SNFEN has just 4.9M parameters, which makes it easily deployable on mobile devices, and being implemented in Keras with TensorFlow further facilitates its optimization for on-device inference.

The source code for both GBFEN and SNFEN is publicly accessible on GitHub, enabling researchers and developers to replicate the experiments and explore potential improvements: https://github.com/raffaele-cappelli/pyfing.

## C. METHODS EVALUATED

In the following sections, the two proposed methods are compared with 11 prominent fingerprint enhancement techniques from the literature. Table 3 presents, for each method, the reference, the year of publication, and a concise description. The methods include frequency-domain approaches (SpectralDict, PCF, and SFP), dictionary-based techniques (GlobalDict, LocalDict, RidgeDict, and SpectralDict), and deep learning-based methods (FingerNet, DenseUNet, Autoenc, PCF, DNUNets, FingerGAN, and SFP).

## D. VISUAL COMPARISON

Figure 9 illustrates how three portions of latent fingerprints from NIST SD27 (one "Good", one "Bad", and one "Ugly") are enhanced by each method listed in Table 3.

Fingerprint (a) exhibits a fairly distinguishable ridge-valley pattern, except for a horizontal band where the pattern is almost completely obscured by noise. All methods successfully identify the general structure of the ridge lines, although the results of some are marred by more or less visible artifacts due to the subdivision of the image into square blocks (see, for example, the results of SFP or FingerNet). In the central noise band, almost all methods encounter difficulties: the noise band remains present, or the corresponding ridge-line pattern is reconstructed unrealistically (as in DNUNets and FingerGAN). Only DenseUNet and SNFEN show some ability to mitigate the impact of this horizontal noise band.

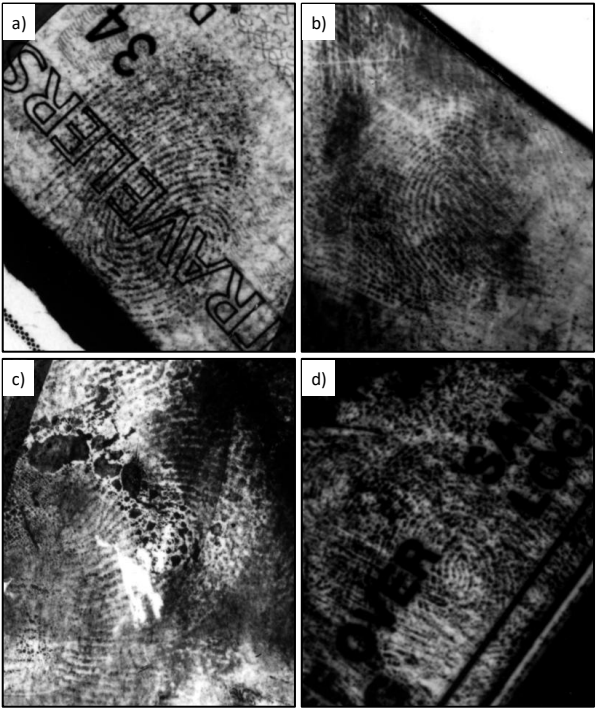

**FIGURE 8.** Sample images from NIST SD27: (a) "Good" fingerprint, (b) "Bad" fingerprint, and (c, d) two "Ugly" fingerprints.

**TABLE 3.** Summary of the state-of-the-art methods considered and the two new proposed ones.

| Method | Year | Description |
|---|---|---|
| GlobalDict [34] | 2013 | Initial orientation estimates are refined using a learned dictionary of reference orientation patches and compatibility constraints; contextual Gabor filtering produces the final enhanced image |
| LocalDict [35] | 2014 | Localized dictionaries refine initial orientation estimates based on fingerprint pose and learned spatial distributions of prototype orientation patches; contextual Gabor filtering produces the final enhanced image |
| RidgeDict [36] | 2014 | Cartoon-texture decomposition removes structured noise, while a ridge-line patch dictionary recovers ridge information; orientation and frequency features, extracted from reconstructed patches, guide contextual Gabor-based enhancement |
| FingerNet [39] | 2017 | Unified fingerprint processing framework that integrates traditional fingerprint processing steps into a single neural network that includes orientation field estimation and fingerprint enhancement |
| SpectralDict [12] | 2018 | After a cartoon-texture decomposition preprocessing step, a spectral dictionary learns filters in the frequency domain to enhance the fingerprint, attempting to preserve curved ridges and minutiae |
| DenseUNet [19] | 2019 | A CNN, trained on noisy fingerprint data, iteratively enhances fingerprint patches by learning complex mappings from low-quality to high-quality images |
| Autoenc [43] | 2020 | A convolutional autoencoder enhances the fingerprint by learning from degraded high-quality images, focusing on the texture component obtained from cartoon-texture decomposition to remove structured noise |
| PCF [21] | 2021 | A progressive and corrective feedback framework enhances the fingerprint using boosted spectral filtering and a spectral autoencoder to detect and correct errors, after a cartoon-texture decomposition preprocessing step |
| DNUNets [45] | 2021 | A Deep Nested U-Net architecture, trained from synthetically-generated latent fingerprints, enhances the fingerprint by learning complex mappings from low-quality to high-quality images |
| FingerGAN [46] | 2023 | A generative adversarial network enhances the fingerprint by generating realistic fingerprint images conditioned on the fingerprint valley-skeleton and a global orientation model, after a cartoon-texture decomposition preprocessing step |
| SFP [22] | 2024 | A spectral filter predictor network iteratively enhances the fingerprint in the frequency domain, after cartoon-texture decomposition, prioritizing high-quality regions and leveraging spectral information to improve low-quality areas |
| GBFEN | 2025 | The proposed contextual Gabor convolution-based method |
| SNFEN | 2025 | The proposed deep learning-based method |

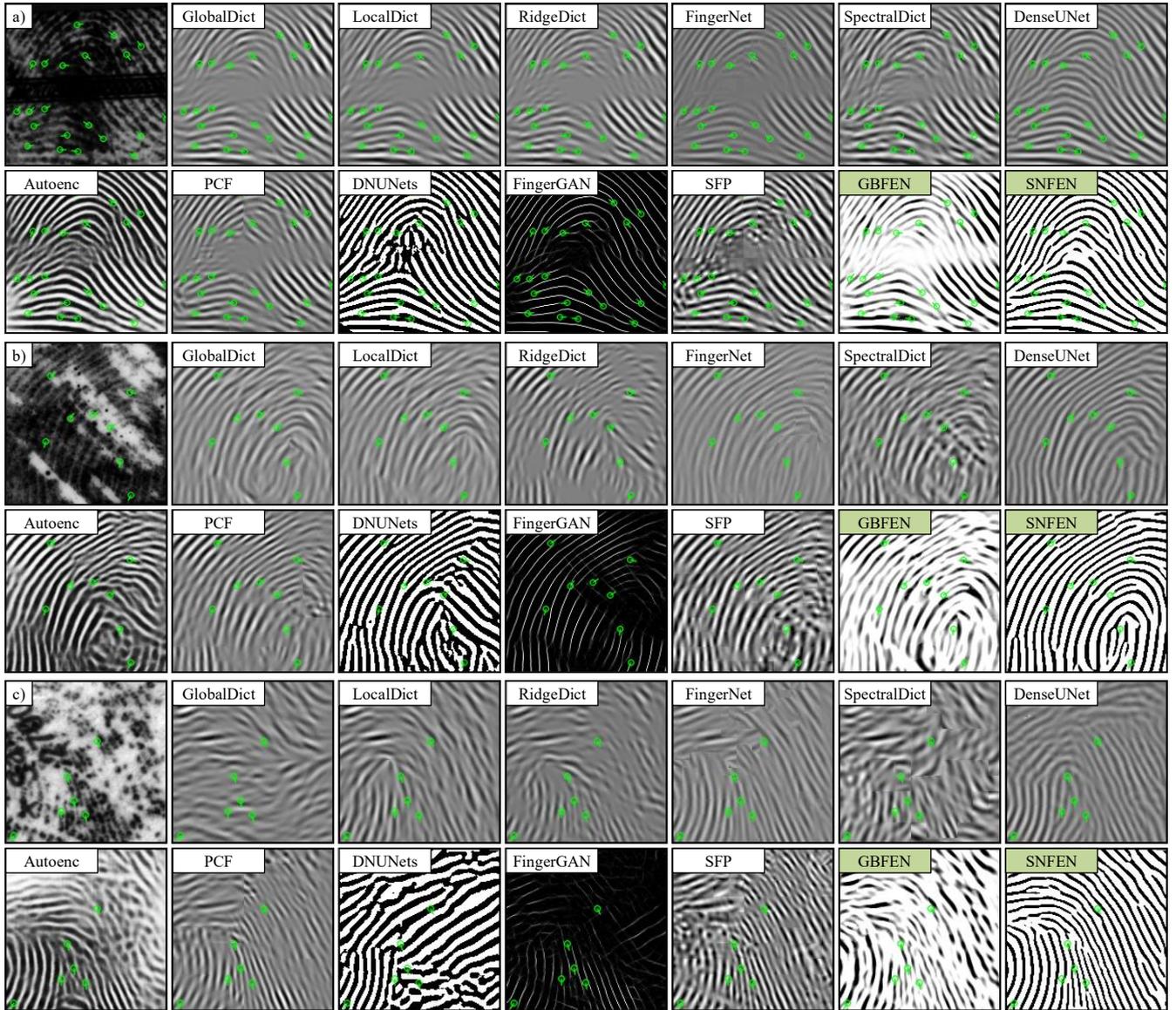

**FIGURE 9.** Visual comparison of fingerprint enhancement results on three portions of fingerprints from NIST SD27: (a) portion of a "Good" fingerprint, (b) portion of a "Bad" fingerprint, and (c) portion of a "Ugly" fingerprint. Note that GBFEN and SNFEN are designed to produce results with white ridges and black valleys. To facilitate comparison with other methods, which use black ridges and white valleys, their enhanced images are presented as negative images in this figure. Ground truth minutiae from NIST SD27 (see section V.E) are reported on all images.

Additionally, it can be observed that most methods produce a low-contrast grayscale result; exceptions include DNUNets, GBFEN, SNFEN, and FingerGAN: the first three produce highly contrasted, almost binary images, while the fourth produces the skeleton of the valleys.

Fingerprint (b) in Figure 9 contains a significant amount of noise, primarily in the form of wide, light, diagonally oriented stripes. It can be observed that the noise negatively influences many methods, which tend to misestimate the orientation of the ridge lines in the right part of the image. Among the methods capable of better estimating orientations in this fingerprint, the resulting images remain quite confused and low-contrast. Exceptions to this include GBFEN and, especially, SNFEN, which yield significantly more accurate results.

Fingerprint (c) in Figure 9 is the most challenging of the three: various types of noise make the ridge pattern difficult to distinguish even for the human eye. All methods fail to estimate the structure of the ridge lines, except for the two proposed methods GBFEN and SNFEN. GBFEN, however, produces an image with very discontinuous ridges, while SNFEN obtains a significantly better result.

In general, from the analysis of Figure 9, the advantage of GBFEN and SNFEN over state-of-the-art methods is evident: the produced images appear more accurate in orientations, clearer, more contrasted, and less influenced by noise. Furthermore, between the two, SNFEN obtains much more continuous and homogeneous ridges. These results are particularly impressive given the simplicity of the proposed methods. Unlike many state-of-the-art techniques, GBFEN and SNFEN do not require complex preprocessing steps (such as cartoon-texture decomposition), iterative optimization, image block division, frequency domain operations, or complicated loss functions. Furthermore, they have not been specifically trained on latent fingerprints, relying solely on a small dataset of live-scanned fingerprints, see also section V.G.

To further validate the quality of the results obtained, two human experts were asked to evaluate the enhancement of ten fingerprints from the "Ugly" subset of the NIST SD27. The experts evaluated the images, enhanced using each method listed in Table 3, and selected the top three for each fingerprint, akin to awarding gold, silver, and bronze medals. The results are summarized in Table 4, which shows the percentage of times each enhancement method was ranked first, second, and third. The methods are listed in ascending order of effectiveness, with the best-performing method at the bottom. Once again, it can be observed that SNFEN achieves by far the best results according to the judgment of the two experts, followed by GBFEN.

TABLE 4. Method ranking based on expert evaluation of ten "Ugly" latent fingerprints from NIST SD27. Best performance at the bottom.

| Method | Ranked first (%) | Ranked second (%) | Ranked third (%) |
|---|---|---|---|
| FingerGAN | 0 | 0 | 0 |
| PCF | 0 | 0 | 0 |
| RidgeDict | 0 | 0 | 0 |
| SpectralDict | 0 | 0 | 0 |
| FingerNet | 0 | 0 | 5 |
| SFP | 0 | 0 | 5 |
| DenseUNet | 0 | 0 | 15 |
| LocalDict | 0 | 0 | 15 |
| Autoenc | 0 | 10 | 15 |
| GlobalDict | 5 | 0 | 0 |
| DNUNets | 5 | 5 | 20 |
| GBFEN | 10 | 65 | 25 |
| SNFEN | 80 | 20 | 0 |

### E. EVALUATION METRICS

Given the unavailability of a noise-free ridge/valley pattern ground truth for NIST DB27 fingerprints, and considering the unfeasibility of manually annotating such a ground truth due to the intricate noise present, minutiae were employed as a surrogate ground truth. Minutiae are already provided as ground truth in the database, defined by their $(x, y)$ position, direction (an angle between zero and $2\pi$), and type (ridge ending or bifurcation).

To evaluate the performance of each enhancement method, minutiae were extracted from each enhanced image **E** using the commercial VeriFinger SDK v13.1 [62]. These extracted minutiae were then compared to the ground truth, considering only those within[3] the ground truth segmentation mask **S**. Standard object detection evaluation metrics were adopted to assess the enhancement method's performance. Specifically, the following measures were calculated over the whole database.

- *True Positives* ($TP$): The total number of extracted minutiae that correspond to ground truth minutiae. Note that various criteria can be defined to determine minutiae correspondence, which will be discussed in detail later.
- *False Positives* ($FP$): The total number of extracted minutiae that do not correspond to ground truth minutiae. These represent spurious minutiae generated by the combined enhancement and extraction process, which can potentially mislead examiners.
- *False Negatives* ($FN$): The total number of ground truth minutiae that are not found among the extracted minutiae. This indicates true features that were either obscured by the enhancement or missed by the subsequent extraction process.

From these values, the following metrics were computed:

- $Precision = \frac{TP}{TP+FP}$: The proportion of correct minutiae among all detected minutiae. High precision is essential as it quantifies the system's ability to minimize the risk of producing spurious minutiae ($FP$).
- $Recall = \frac{TP}{TP+FN}$: The proportion of ground truth minutiae that were detected.
- $F_1\text{-}score = 2 \cdot \frac{Precision \cdot Recall}{Precision+Recall} = \frac{2 \cdot TP}{2 \cdot TP+FP+FN}$: The harmonic mean of $Precision$ and $Recall$, also known as Dice coefficient [63], providing a balanced measure of the performance [64].

To determine the correspondence between two minutiae, the following elements were considered: Euclidean distance between their positions (required to be less than or equal to a threshold $\tau_D$), difference in their directions (required to be less than or equal to an angle $\tau_\theta$), and minutiae type. Regarding minutiae type, two approaches were adopted: (1) requiring an exact match between the types of corresponding minutiae, as in [46], and (2) ignoring the type, acknowledging that an ambiguity may exist between ridge ending and bifurcation minutiae types [1]. Results are presented for $\tau_D = 14$ and $\tau_\theta = \frac{\pi}{9}$, considering both exact type matching and type-agnostic matching. Additionally, results for varying $\tau_D$ and $\tau_\theta$ thresholds are provided to offer a more comprehensive analysis.

### F. QUANTITATIVE COMPARISON

This section presents and analyzes the quantitative comparison results among the considered methods, using the metrics defined in the previous section.

Table 55 reports the results obtained by comparing the extracted minutiae with the ground truth minutiae, imposing the same minutiae type. SNFEN and GBFEN achieve the best results, with the highest and second-highest $F_1$-$score$, respectively. SNFEN's result is due to its high *Precision*, the highest among all methods, combined with a very good *Recall* value, although not the highest. Conversely, GBFEN excels in *Recall*, achieving the highest value, with the second-highest *Precision* value.

---

[3] Only minutiae located at a minimum distance of 14 pixels from the boundary of the ground truth segmentation mask were considered.

Table 66 presents the results obtained by comparing the extracted minutiae with the ground truth minutiae without imposing the same minutiae type. Again, SNFEN and GBFEN achieve the first and second place in the ranking according to the $F_1$-*score*, demonstrating their ability to produce the best enhanced images.

**TABLE 5.** Method ranking by $F_1$-*score*, using $\tau_D = 14$, $\tau_\theta = \frac{\pi}{9}$, and exact minutia type matching. Best results in bold, best performance at the bottom.

| Method | Precision | Recall | $F_1$-score |
|---|---|---|---|
| GlobalDict | 0.150 | 0.380 | 0.215 |
| Autoenc | 0.153 | 0.380 | 0.218 |
| FingerGAN | 0.165 | 0.395 | 0.233 |
| LocalDict | 0.168 | 0.405 | 0.238 |
| FingerNet | 0.165 | 0.431 | 0.238 |
| SpectralDict | 0.163 | 0.465 | 0.242 |
| RidgeDict | 0.175 | 0.395 | 0.242 |
| PCF | 0.173 | 0.429 | 0.247 |
| DenseUNet | 0.186 | 0.392 | 0.252 |
| DNUNets | 0.198 | 0.389 | 0.263 |
| SFP | 0.187 | 0.452 | 0.264 |
| GBFEN | 0.201 | **0.495** | 0.286 |
| SNFEN | **0.236** | 0.436 | **0.306** |

Table 77 provides a detailed breakdown of the results for the two proposed methods in the case of exact minutia type matching (as in Table 55), for each of the three fingerprint subsets ("Good", "Bad", "Ugly") that comprise the test dataset. As expected, the performance is highest in the "Good" set, decreases in the "Bad" set, and further declines in the "Ugly" set. However, the performance trend of the two methods remains consistent across the three subsets.

For the reader's convenience, the table also includes the results for the entire dataset (as in Table 55).

**TABLE 6.** Method ranking by $F_1$-*score*, using $\tau_D = 14$, $\tau_\theta = \frac{\pi}{9}$, and type-agnostic minutia matching. Best results in bold, best performance at the bottom.

| Method | Precision | Recall | $F_1$-score |
|---|---|---|---|
| GlobalDict | 0.269 | 0.681 | 0.385 |
| Autoenc | 0.276 | 0.688 | 0.394 |
| SpectralDict | 0.267 | 0.759 | 0.395 |
| LocalDict | 0.289 | 0.694 | 0.408 |
| FingerGAN | 0.291 | 0.697 | 0.411 |
| FingerNet | 0.285 | 0.745 | 0.412 |
| RidgeDict | 0.305 | 0.690 | 0.423 |
| PCF | 0.300 | 0.742 | 0.427 |
| DenseUNet | 0.320 | 0.674 | 0.434 |
| SFP | 0.317 | 0.768 | 0.449 |
| DNUNets | 0.340 | 0.666 | 0.450 |
| GBFEN | 0.322 | **0.791** | 0.457 |
| SNFEN | **0.390** | 0.720 | **0.506** |

The graphs in Figure 10 confirm the results of the two proposed methods. SNFEN clearly dominates all other methods for any reasonable value of $\tau_D$ and $\tau_\theta$, both considering the minutiae type and ignoring it. GBFEN also generally achieves better results than other methods, but with a smaller margin in the case of type-agnostic minutiae matching. Only one method (DNUNets) reaches and slightly surpasses GBFEN for very high values of threshold $\tau_D$.

As previously noted, these results, which are excellent in absolute terms, become exceptional when interpreted in light of the vast difference in complexity between the proposed methods and the most accurate state-of-the-art methods, see section V.G.

**TABLE 7.** Detailed performance metrics of SNFEN and GBFEN for exact minutia type matching, across the three different fingerprint quality groups ("Good", "Bad", "Ugly") and the entire test dataset.

| | All fingerprints | | | "Good" fingerprints | | | "Bad" fingerprints | | | "Ugly" fingerprints | | |
|---|---|---|---|---|---|---|---|---|---|---|---|---|
| Method | Precision | Recall | $F_1$-score | Precision | Recall | $F_1$-score | Precision | Recall | $F_1$-score | Precision | Recall | $F_1$-score |
| GBFEN | 0.201 | 0.495 | 0.286 | 0.304 | 0.511 | 0.381 | 0.162 | 0.485 | 0.243 | 0.123 | 0.465 | 0.195 |
| SNFEN | 0.236 | 0.436 | 0.306 | 0.341 | 0.460 | 0.392 | 0.195 | 0.426 | 0.267 | 0.143 | 0.383 | 0.209 |

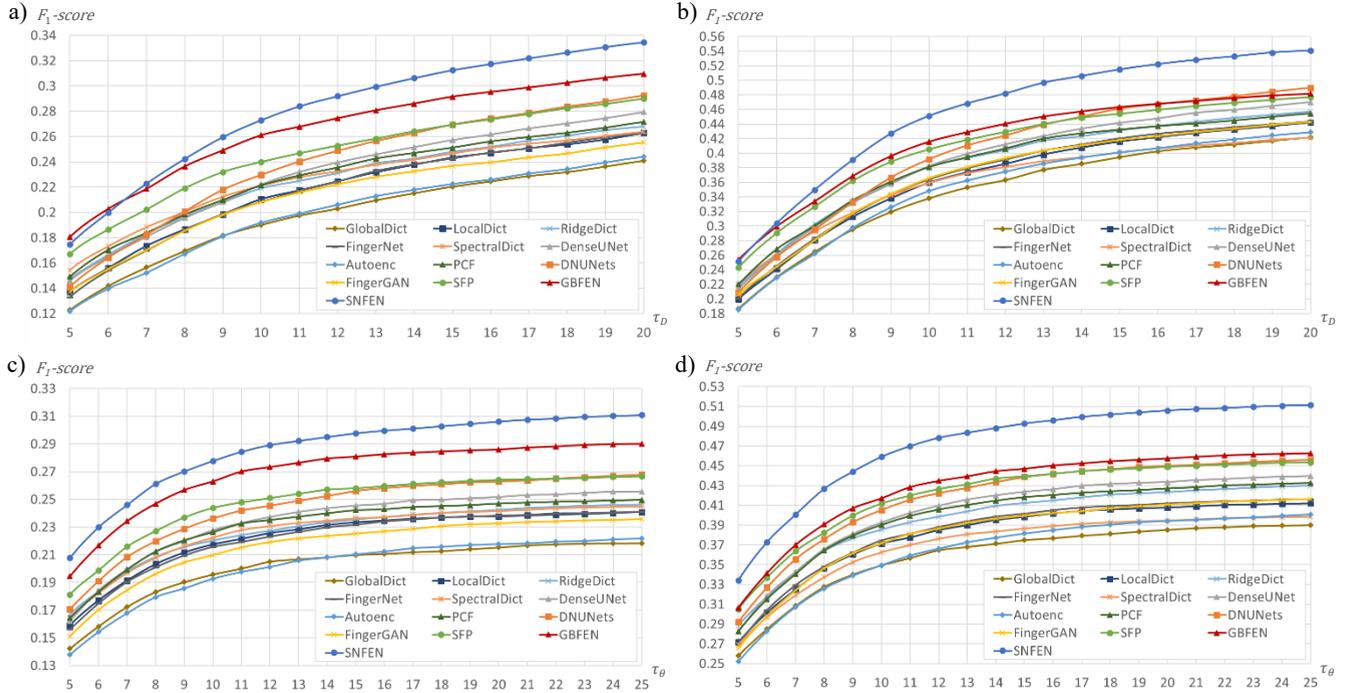

**FIGURE 10.** Top row: $F_1$-*score* as a function of $\tau_D$ (in pixels), with $\tau_\theta = \frac{\pi}{9}$, and (a) exact minutia type matching or (b) type-agnostic minutia matching. Bottom row: $F_1$-*score* as a function of $\tau_\theta$ (in degrees, between 5° and 25°), with $\tau_D = 14$, and (c) exact minutia type matching or (d) type-agnostic minutia matching.

### G. ABOUT COMPLEXITY

While the analysis of the outcomes from various enhancement methods is of crucial importance, it is equally necessary to consider their complexity to fully evaluate their strengths and weaknesses. From this perspective, GBFEN becomes particularly noteworthy, as it is undoubtedly the simplest method among those considered. SNFEN, although undeniably more complex than GBFEN, remains relatively simple when compared to state-of-the-art methods based on neural networks. Consider the data presented in Table 88, which compares SNFEN with three recent state-of-the-art methods based on the following metrics:

- *Network architecture* – The simplicity of the network architecture directly correlates with the ease of implementation and computational overhead.
- *Loss function(s)* – The choice of loss function significantly impacts the simplicity of a method. A straightforward, well-understood loss function allows for easier optimization and interpretation of results, making the method simpler to implement and troubleshoot.
- *Training data* – Evaluating simplicity involves assessing the dataset used for training. Methods employing fewer fingerprint images for training and naiver data augmentation procedures can be deemed simpler due to reduced data preprocessing requirements and less computational resource demand.
- *Training time* – Training time serves as a direct indicator of method complexity. A technique that achieves comparable results with significantly less training time exemplifies simplicity by reducing resource consumption and enabling quicker deployment.
- *Fingerprint enhancement procedure* – The procedural steps in fingerprint enhancement dictate the method's complexity. A procedure with fewer steps or one that integrates preprocessing and enhancement in a streamlined manner is inherently simpler, reducing the likelihood of errors and simplifying the workflow.
- *Open source* – Open-source availability is pivotal for assessing simplicity. A method that is open-source allows for community review, which can lead to simplifications through community-driven optimizations and bug fixes. Moreover, it provides transparency about the method's intricacies, making it easier for others to understand, modify, or integrate into existing systems without proprietary barriers, thereby enhancing perceived and actual simplicity.

Analyzing the information presented in Table 88, in light of the importance of the elements described above, the significant difference in complexity is more than evident.

TABLE 8. Comparison of complexity metrics.

| Metric | SNFEN | DNUNets [45] | FingerGAN [46] | SFP [22] |
|---|---|---|---|---|
| Network architecture | Encoder-decoder with skip connections | A complex structure consisting of an encoder with multiple decoders nested at each encoding level, with several skip connections originating both from encoding and decoding blocks | A Generator network (Encoder-decoder with skip connection), and a Discriminator network (a classical CNN). The two networks form a GAN | Encoder-decoder with skip connections |
| Loss function(s) | A simple loss based on the Tversky index | Weighted sum of two loss functions: mean squared error and local structural similarity index [65] | An Adversarial Loss to jointly train the discriminator and the generator, and a Reconstruction Loss, to optimize minutia information when training the generator | A simple loss function based on the cosine similarity |
| Training data | 360 fingerprints with simple data augmentation | 30.000 latent fingerprints synthetically generated from real fingerprints from NIST SD14 [66] adding various noise backgrounds, including backgrounds of NIST SD27 images | 130.000 latent fingerprints synthetically generated from 13.000 real fingerprints from NIST SD14 [66] adding noise backgrounds of NIST SD27 images | 29.700 fingerprints from NIST SD14 combined with backgrounds from the Describable Textures Dataset [67] |
| Training time | 25 minutes | 24 hours | Not reported | 2.5 hours |
| Fingerprint enhancement procedure | A single network inference | A single network inference | After a cartoon-texture decomposition preprocessing, the image is divided into patches of 192 × 192 pixels, a network inference is executed for each patch, then the enhanced patches are combined to obtain the final enhanced image | After a cartoon-texture decomposition preprocessing and the selection of initial high-quality patches, a complex iterative procedure progressively enhances the fingerprint in the frequency domain, using the deep network to predict the spectral filters, prioritizing image patches based on quality, leveraging the improved spectra from previously enhanced patches to enhance lower-quality ones. |
| Open source | Yes | No | Yes | No |

## H. ABLATION STUDY

As a final experiment to validate SNFEN's simplicity and the essentiality of its components, several ablation tests were performed. The fingerprint enhancement network was trained from scratch and evaluated after removing each of the following elements individually:

- The input segmentation mask **S**.
- The input orientation field **Θ**.
- The input frequency map **$\mathcal{F}$**.
- All input features, leaving just the fingerprint **F** as the sole input of the network.
- All skip connections except the first one[4].
- The last level of the Encoder and Decoder blocks.
- The 5 × 5 filters in the convolution blocks (replacing them with 3 × 3 filters).
- Half of the convolution filters in each level of the Encoder and Decoder blocks.
- Data augmentation on the training fingerprints.

Table 99 shows the results of this ablation study. All evaluated networks exhibited performance degradation, confirming that all the considered components are essential, and the proposed method adopts the simplest possible configuration with optimal performance.

TABLE 9. Results of the ablation study, with $\tau_D = 14$, $\tau_\theta = \frac{\pi}{9}$, and exact minutia type matching.

| Element removed | $F_1$-score |
|---|---|
| None | 0.306 |
| Input segmentation mask **S** | 0.299 |
| Input orientation field **Θ** | 0.296 |
| Input frequency map **$\mathcal{F}$** | 0.287 |
| All input features except the fingerprint **F** | 0.285 |
| All skip connections but the first one | 0.273 |
| Last level of the Encoder and Decoder blocks | 0.302 |
| 5 × 5 convolution filters (replaced by 3 × 3 filters) | 0.298 |
| Half convolution filters in each Encoder and Decoder level | 0.297 |
| Training data augmentation | 0.282 |

---

[4] It was not possible to remove all skip connections, as this prevented the training process from converging.

## VI. CONCLUSION

This study demonstrates that simplicity can be a powerful tool in fingerprint enhancement, achieving state-of-the-art performance with minimalist approaches. The proposed methods, GBFEN and SNFEN, leverage simple techniques — contextual convolution with Gabor filters and an Encoder-Decoder architecture with skip connections, respectively — to effectively enhance low-quality latent fingerprints, starting from local orientation and frequency features estimated by recently-proposed techniques [49] [50]. The success of this straightforward strategy underscores the critical role of orientation and frequency features in achieving high-quality fingerprint enhancement. While more complex architectures may offer potential for further improvement, the impressive results achieved with these simple methods highlight the need for a balanced approach between complexity and practical benefits.

Furthermore, the open-source implementation of these methods promotes reproducibility and provides a valuable resource for the research community. Publicly accessible code encourages further exploration and advancement in the field.

The two proposed methods exhibit interesting complementarities. GBFEN boasts minimal hardware requirements, necessitating no GPU, making it highly accessible. SNFEN, while requiring GPU training, obviously demonstrates remarkable efficiency with a GPU, as is to be expected, but the network is small enough that it can also be run on a CPU. In any case, this disparity in hardware needs offers a practical advantage depending on the resources at hand. Furthermore, GBFEN excels in *Recall*, at the cost of lower *Precision*, while SNFEN exhibits the opposite behavior. This trade-off between *Recall* and *Precision* allows for method selection based on the specific application requirements.

Both methods, despite being trained exclusively on non-latent fingerprints, demonstrate remarkable generalization capabilities. Their performance on latent fingerprints, which exhibit noise characteristics substantially different from the training data, surpasses current state-of-the-art results, thus underscoring the robustness of the proposed approaches. Crucially, this study avoided the practice, employed by certain previous leading methods (e.g., [45] and [46], as detailed in Table 8), of training on latent fingerprints using noise backgrounds directly derived from the test images. This practice was avoided because it risks inducing overfitting or information leakage from the test set into the training process, potentially undermining real-world applicability. While the proposed methods achieve strong performance, certain limitations warrant consideration. An error analysis across various subsets of the SD27 dataset indicates that performance remains suboptimal on the "Bad" and "Ugly" fingerprint categories. This highlights a persistent challenge in effectively processing severely degraded prints (e.g., Figure 11) and suggests avenues for future work. For instance, exploring alternative training strategies, incorporating latent fingerprints into the training data, or adopting novel forms of data augmentation could potentially improve performance, and investigating alternative network architectures might lead to further advancements in this specialized field.

Additionally, while effective on live-scanned and latent fingerprints, the methods' efficacy on drastically different acquisition techniques, such as touchless fingerprint acquisition [1], requires investigation. It is also important to note that both methods are trained at a fixed resolution of 500 dpi. Consequently, they may not directly handle different resolutions without image rescaling, which could impact performance.

In conclusion, this research reaffirms the value of the KISS principle in modern fingerprint recognition systems. The findings suggest that future research should prioritize a balance between complexity and performance, carefully exploring simple yet effective solutions. Future work will apply a similar strategy to other fingerprint recognition processing steps, such as minutiae extraction.

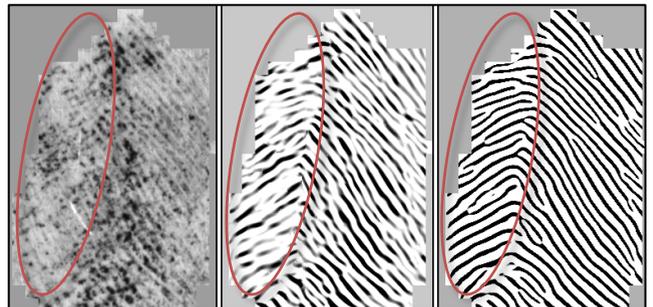

**FIGURE 11.** Example of an "Ugly" fingerprint from NIST DB27 where both GBFEN and SNFEN fail to correctly interpret the ridge orientation in a consistent portion of the print (circled in all three images).

## ACKNOWLEDGMENT
The author would like to express sincere gratitude to Dr. Manhua Liu, Qian Peng, and Supakit Kriangkhajorn for generously providing enhanced fingerprint images obtained using various methods. Additionally, the author thanks Dr. Jiankun Hu, Dr. Yanming Zhu, and Dr. Xuefei Yin for sharing the source code and experimental details of their method. The author also extends heartfelt thanks to Dr. Matteo Ferrara and Dr. Davide Maltoni for their invaluable advice and expertise in fingerprint analysis.